\documentclass[10pt,twocolumn,letterpaper]{article}

\usepackage{cvpr}
\usepackage{times}
\usepackage{epsfig}
\usepackage{graphicx}
\usepackage{amsmath}
\usepackage{amssymb}
\usepackage{diagbox}
\usepackage{color}

\usepackage{graphicx}
\usepackage{subfigure}

\usepackage[breaklinks=true,bookmarks=false]{hyperref}

\cvprfinalcopy 

\def\cvprPaperID{****} 

\begin{document}

\title{Semantic Alignment: Finding {Semantically} Consistent {Ground-truth} for Facial Landmark Detection}

\author{\normalsize{Zhiwei Liu$^{1,2}$}\thanks{equal contribution.}, \normalsize{Xiangyu Zhu$^{1*}$}, 
	\normalsize{Guosheng Hu$^{4,5}$}, \normalsize{Haiyun Guo$^{1}$}, 
	\normalsize{Ming Tang$^{1,3}$}, \normalsize{Zhen Lei$^{1}$}, 
	\normalsize{Neil M. Robertson$^{5,4}$ and Jinqiao Wang$^{1,3}$}\\
	\small{$^{1}$National Lab of Pattern Recognition, Institute of Automation, CAS, Beijing 100190, China} 
	\small{$^{2}$University of Chinese} \\ \small{Academy of Sciences}  
	\small{$^{3}$Visionfinity Inc., ObjectEye Inc., Universal AI Inc.}
	\small{$^{4}$AnyVision} 
	\small{$^{5}$Queen’s University Belfast}\\
	{\small \tt $^{1}${\{zhiwei.liu, xiangyu.zhu, haiyun.guo, tangm, zlei, jqwang\}}@nlpr.ia.ac.cn} \\  
	{\small \tt $^{4}${huguosheng100@gmail.com}}  \ \ {\small\tt $^{5}${N.Robertson@qub.ac.uk}}
}
\maketitle

\begin{abstract}
	Recently, deep learning based facial landmark detection has achieved great success. Despite this, we notice that 
	the semantic ambiguity greatly degrades the detection performance. Specifically, the semantic ambiguity means 
	that some landmarks (e.g. those evenly distributed along the face contour) do not have clear and accurate definition, 
	causing inconsistent annotations by annotators. Accordingly, 
	these inconsistent annotations, which are usually provided by public databases, commonly work as the  
	ground-truth to supervise network training, leading to the degraded accuracy.
	To our knowledge, little research has investigated this problem. 
	In this paper, we propose a novel probabilistic model which introduces a latent variable, i.e. the `real' ground-truth which 
	is semantically consistent, to optimize. This framework couples two parts (1) training landmark detection CNN and (2) 
	searching the `real' ground-truth. These two parts are alternatively optimized: the searched `real' ground-truth 
	supervises the CNN training; and the trained CNN assists the searching of `real' ground-truth. 
	In addition, to recover the unconfidently predicted landmarks due to occlusion and low quality, 
	we propose a global heatmap correction unit (GHCU) to correct outliers by considering the global face shape as a constraint. 
	Extensive experiments on both image-based (300W and AFLW) and video-based (300-VW) databases demonstrate that our 
	method effectively improves the landmark detection accuracy and achieves the state of the art performance.
\end{abstract}

\section{Introduction}

\begin{figure}[t]
	\begin{center}
		\includegraphics[width=\linewidth]{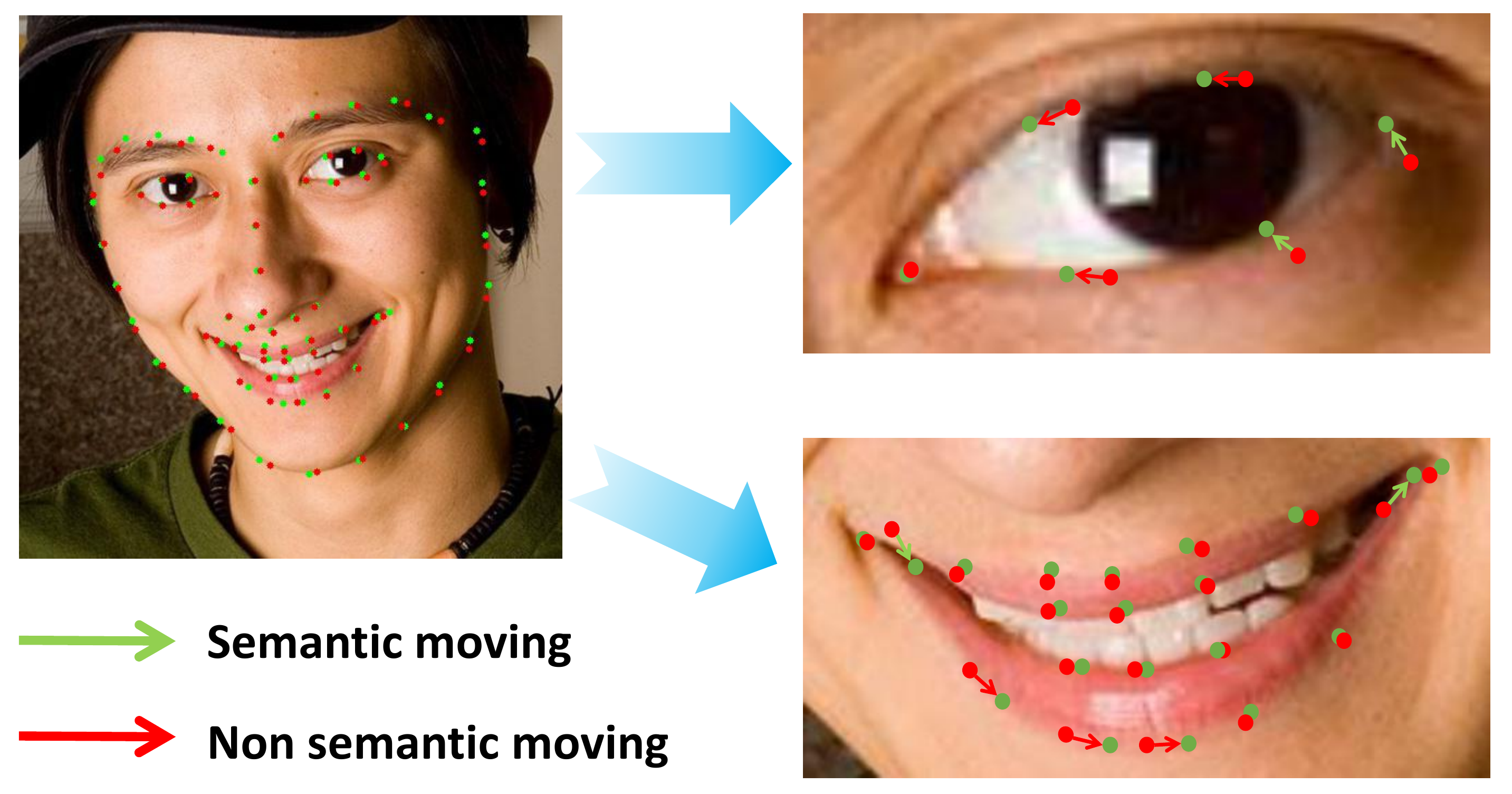}
	\end{center}
	\caption{ {The  landmark updates in training after the model is roughly converged. Due to `semantic ambiguity', we can see that 
			many optimization directions, which are random guided by random annotation noises {along with the contour and 
				`non semantic'.} The others move to the semantically accurate positions. 
			Red and green dots denote the predicted and annotation landmarks, respectively.
	}}
	\label{fig:problem}
\end{figure}

Deep learning methods \cite{Sun2013Deep, zhang2016learning, zhu2017face, lv2017deep, feng2017wing, wu2018look} have achieved 
great success on landmark detection due to the strong modeling capacity.
Despite this success, precise and credible landmark detection still has many challenges, one of which  is the degraded performance caused by `semantic ambiguity'. This ambiguity results from the lack of clear definition  on those weak semantic landmarks {on the contours} (e.g. those on face contour and nose bridge). In comparison, strong semantic landmarks on the corners (e.g. eye corner) {suffer less from such 
	ambiguity}. 
The `semantic ambiguity' can make human annotators confused about the positions of weak semantic points, and it is inevitable for annotators to introduce random noises {during annotating}. The inconsistent and imprecise annotations can mislead CNN training and cause degraded performance. 
Specifically, when the deep model roughly converges to the ground-truth provided by public databases, the network training is misguided by random annotation noises caused by `semantic ambiguity', shown in Fig.~\ref{fig:problem}. Clearly these noises can make the network training trapped into local minima, leading to degraded results. 

In this paper, we propose a novel Semantic Alignment method which reduces the `semantic ambiguity'  
intrinsically. We assume that there exist  `real' ground-truths which are semantically
consistent and more accurate than human annotations provided by databases.
We model the `real' ground-truth as a latent variable to optimize, and the optimized `real' ground-truth then supervises the landmark detection network training. 
Accordingly, we propose a probabilistic model which can simultaneously search the `real' ground-truth and train the landmark detection network in an end-to-end way. 
In this  probabilistic model, {the \emph{prior model} is to constrain the latent variable to be close to the observations of the `real' ground truth, one of which is the human annotation.} 
The \emph{likelihood model} is to reduce the Pearson Chi-square distance between the expected and the predicted distributions of `real' ground-truth. 
{The heatmap generated by the hourglass architecture \cite{newell2016stacked} represents the confidence 
	of each pixel and this confidence distribution is used to model the predicted distribution of likelihood.}
Apart from the proposed probabilistic framework, {we further propose a global heatmap correction unit (GHCU) which maintains the global face shape constraint and recovers the unconfidently predicted landmarks caused by challenging factors such as occlusions and low resolution of images.}
We conduct experiments on 300W~\cite{Sagonas2013A}, 
AFLW~\cite{koestinger2011annotated} and 300-VW~\cite{shen2015first, tzimiropoulos2015project, chrysos2015offline}
databases and achieve the state of the art performance.

\section{Related work} 

In recent years, convolutional neural networks (CNN) achieves very impressive results on many computer vision tasks including face alignment. Sun $et\ al$~\cite{Sun2013Deep} proposes to cascade several DCNN to predict the shape stage by stage. Zhang $et\ al$~\cite{Zhang2014Facial} proposes a single CNN and jointly optimizes facial landmark detection together with facial attribute recognition,  further enhancing the speed and performance. 
The methods above use 
{shallow}
CNN models to directly regress facial landmarks, which are difficult to cope the complex task with dense landmarks and large pose variations. 

To further improve the performance, many popular semantic segmentation and human pose estimation frameworks are used for face alignment~\cite{Yang2017Stacked,Deng2017Joint, bulat2017far, merget2018robust}. For each landmark, they 
predict a heatmap which contains the probability of  the corresponding landmark. Yang et al. ~\cite{Yang2017Stacked} 
uses a two parts network, i.e., a supervised
transformation to normalize faces and a stacked hourglass network ~\cite{newell2016stacked} to get prediction heatmaps. 
Most recently, JMFA ~\cite{Deng2017Joint} and FAN ~\cite{bulat2017far} also achieve the state of the art accuracy by 
leveraging stacked hourglass network. 
However, these methods do not consider the `semantic ambiguity' problem which potentially degrades the detection performance.

{Two recent works, LAB ~\cite{wu2018look} and SBR ~\cite{dong2018supervision}, are  related to
	this `semantic ambiguity' problem. By introducing more information than pixel intensity only, they implicitly alleviate the impact of the 
	annotation noises and  improve the performance.}
LAB ~\cite{wu2018look} 
trains a facial boundary heatmap estimator and 
incorporates it into 
the main landmark regression network. 
{LAB uses the
	well-defined facial 
	boundaries which 
	{provide}
	the facial geometric structure to}
reduce the ambiguities, leading to improved performance. 
{However, LAB is computational expensive.}
SBR ~\cite{dong2018supervision} proposes a registration loss which 
uses the coherency of optical flow from adjacent frames as its supervision. The additional information from local feature can mitigate the impact of random noises. {However, the optical flow is not always {credible} in unconstrained environment and SBR  trains their model on the testing video before the test, limiting its applications.
} 
{To summarize, LAB and SBR do not intrinsically address the problem of `semantic ambiguity' because the degraded accuracy is actually derived from the inaccurate labels (human annotations provided by databases).
}
In this work, we solve the `semantic ambiguity' problem in a more intrinsic way. 
Specifically, 
we propose a probabilistic model which can simultaneously search the `real' ground-truth without 
semantic ambiguity and train a hourglass landmark detector without using additional information.

\section{Semantic ambiguity}
{
	{The semantic ambiguity indicates that some landmarks do not have clear and accurate definition.}
	In this work, we find 
	the semantic ambiguity can happen on any facial points, but mainly on those weak semantic facial points. {For example, the   landmarks are defined to evenly distribute along the face contour without any clear definition of the exact positions.} This ambiguity can potentially affect: (1) the accuracy of the annotations and (2) the {convergence} of deep model training. For (1), when annotating a database, annotators can introduce random errors to generate inconsistent ground-truths on those weak semantic 
	points due to the lack of clear definitions.
	For (2), the inconsistent ground-truths
	{generate inconsistent}
	gradients for back-propagation, leading to the difficulty of model 
	convergence. In this section, we {qualitatively} analyze the influence of semantic ambiguity on landmark detection. 
}

Before this analysis, we briefly introduce our heatmap-based landmark detection network. Specifically, we use a  four stage Hourglass (HGs) ~\cite{newell2016stacked}. {It can generate the heatmap which provides the 
	probability of the corresponding landmark {located}  
	at every pixel, and this probability can facilitate 
	our analysis of semantic ambiguity.}

\begin{figure}
	\centering
	\subfigure[{The difference between the heatmap of the eye corner (strong semantic) points and the eye 
		contour (weak semantic) points. Col 2 and 3 represent 2D and 3D heatmaps respectively. 
		In the 3D Gaussian, the x, y axes are image coordinates and z axis is the prediction confidence. {We can see 
			the 3D heatmap of a weak semantic point has a `flat hat'.}}]{
		\begin{minipage}[b]{0.45\textwidth}
			\includegraphics[width=\linewidth]{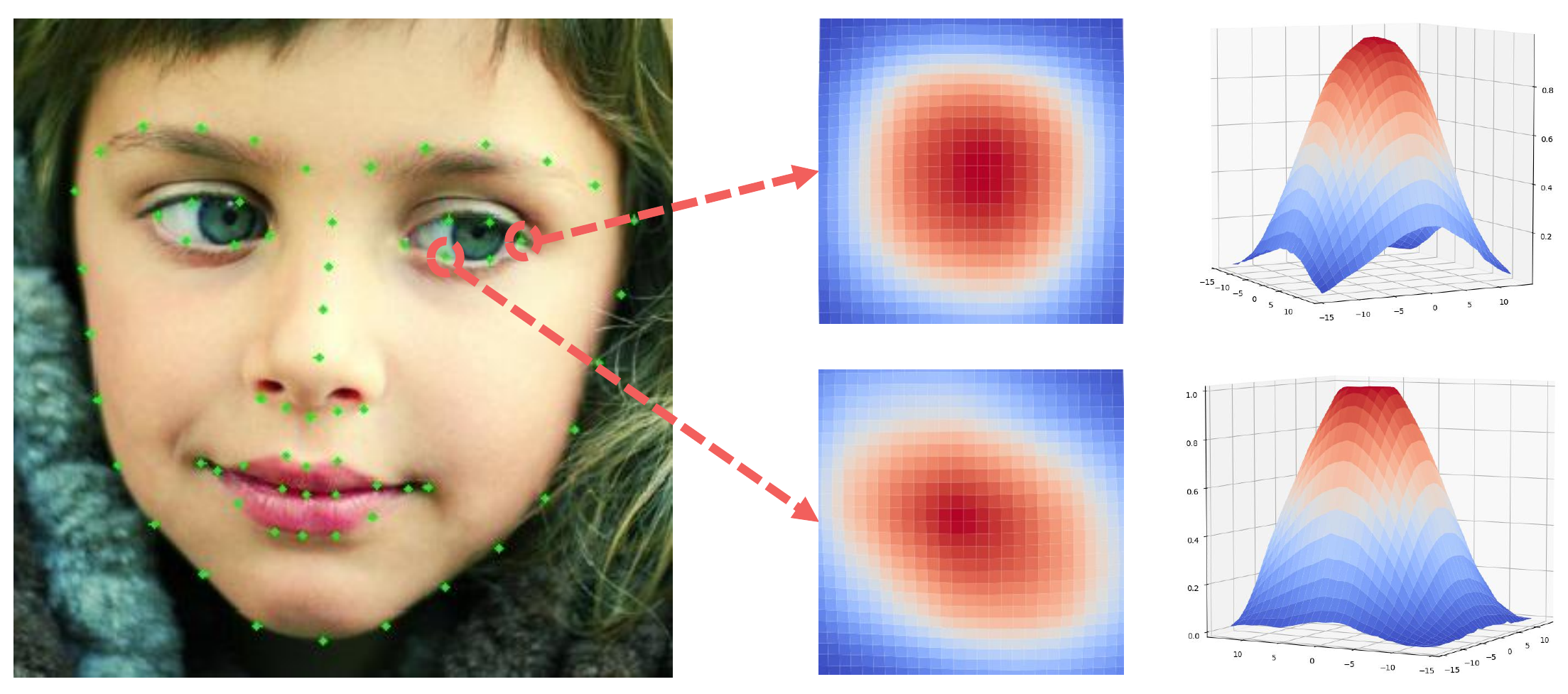} \\
		\end{minipage}
	}
	\subfigure[{ The predictions from a series of checkpoints after convergence.
		When the model has roughly converged, we continue training and achieve the  predictions from different iterations. Red and green dots denote the predicted and annotation landmarks, respectively. {We can see the predicted landmarks from different checkpoints fluctuate in the neighborhood area of the annotated position (green dots).}
	}]{
		\begin{minipage}[b]{0.45\textwidth}
			\includegraphics[width=\linewidth]{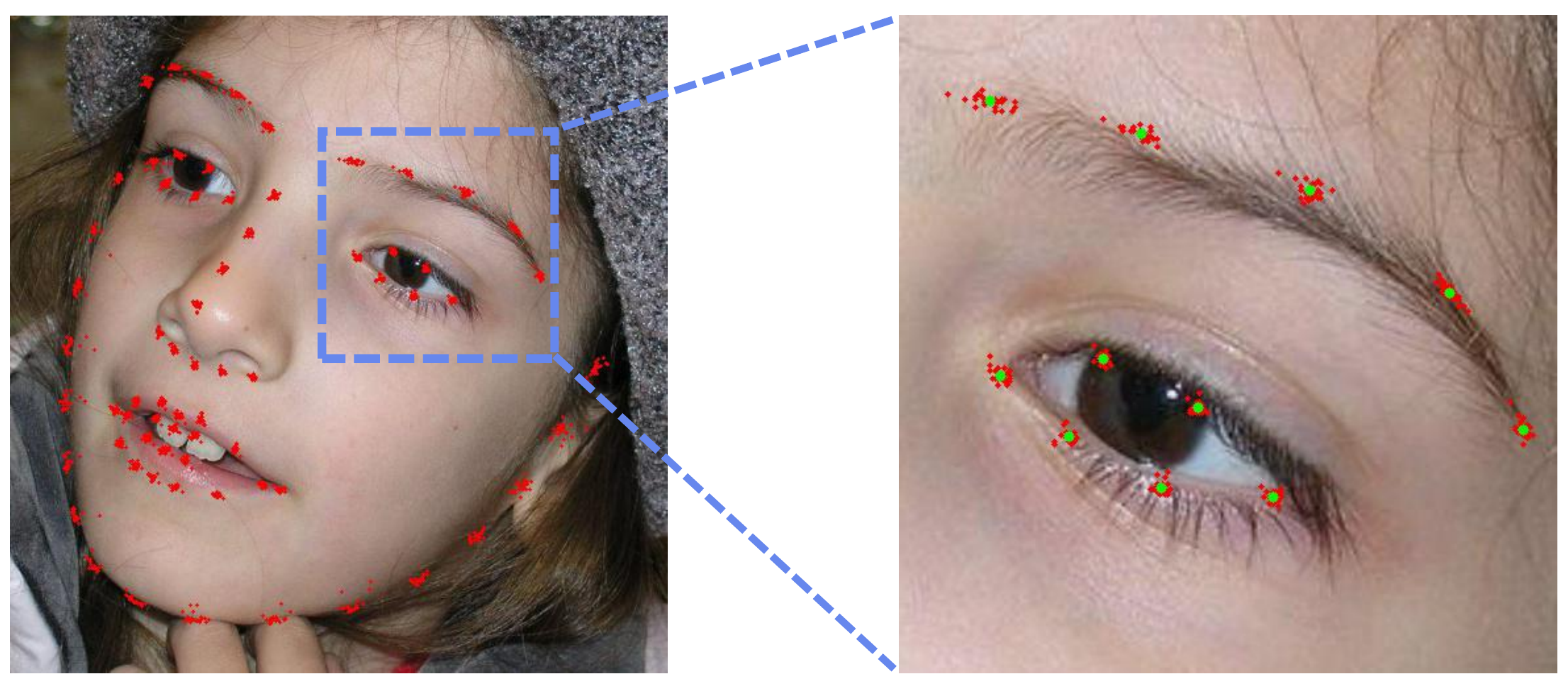}
		\end{minipage}
	}
	\caption{{The effect of semantic ambiguity}}
	\label{fig:analysis}
\end{figure}

Firstly, we find CNN provides a candidate region rather than a confirmed position for a weak semantic point. 
In Fig.~\ref{fig:analysis} (a), we can see that the heatmap of a strong semantic point is nearly Gaussian, 
while the 3D heatmap of a weak semantic point has a `flat hat', meaning that the confidences in that area are very similar. Since 
the position with the highest confidence is chosen as the output. The landmark detector tends to output 
an unexpected random position on the `flat hat'.

{Secondly, we analyze the `semantic ambiguity' by visualizing how the model is optimized after convergence. When the network has roughly converged, we continue training the network and save a series of checkpoints. 
	In Fig.~\ref{fig:analysis} (b), the eyebrow landmarks, from different checkpoints fluctuate along with the 
	edge of eyebrow, which always generates considerable loss to optimize. However, this loss is ineffectual 
	since the predicted points from different checkpoints also fluctuate 
	in the neighborhood area of 
	the annotated position 
	(green dots in Fig.~\ref{fig:analysis} (b)). It can be concluded that the loss caused by random 
	annotation noises dominate the back-propagated gradients after roughly convergence, making the network 
	training trapped into local minima.}

\section{Semantically consistent alignment}
In this section, we detail our methodology. 
In Section 4.1, we model the landmark detection problem using a probabilistic model. 
To deal with the semantic ambiguity caused by human annotation noise, we introduce a latent variable $\hat{\mathbf{y}}$ which represents the `real' ground-truth. Then we model the prior model and likelihood in Section 4.2 and 4.3, respectively.  Section 4.4  proposes an alternative optimization strategy to search $\hat{\mathbf{y}}$ and train the landmark detector. To recover the unconfidently predicted landmarks due 
to occlusion and low quality, we propose a global heatmap correction unit (GHCU) in Section 4.5, 
which refines the predictions by considering the global face shape as a constraint, leading to a more robust model.


\subsection{A probabilistic model of landmark prediction}
In the probabilistic view, training a CNN-based  landmark detector can be formulated as a likelihood 
maximization problem:
\begin{equation}\label{Eq:1}
\max_{\mathbf{W}}\mathcal{L}(\mathbf{W}) = P(\mathbf{o}|\mathbf{x}; \mathbf{W}),
\end{equation}
where  $\mathbf{o} \in \mathbb{R}^{2N}$ is the coordinates of the observation of landmarks (e.g. the human annotations).
$N$ is the number of landmarks, 
$\mathbf{x}$ is the input image and 
$\mathbf{W}$ is the CNN parameters. {Under the probabilistic view of Eq. (\ref{Eq:1}), one pixel value on the heatmap works as the confidence of one particular landmark at that pixel. Therefore, the whole heatmap works}
as the probability distribution over the image. 


{As analyzed in Section 3, the annotations provided by public databases are usually not fully credible due to the  `semantic ambiguity'.
	As a result, the annotations, in particular those of weak semantic landmarks, contain random noises and are inconsistent among faces.
	In this work, we assume that there exists a `real' ground-truth without semantic 
	ambiguity and can better supervise the network training. 
	To achieve this, we introduce a latent variable $\hat{\mathbf{y}}$ as the `real' ground-truth which is optimized 
	during learning.}
{Thus, Eq. (1)  can be reformulated as:}

\begin{equation}\label{Eq:2}
\begin{aligned}
\max_{\hat{\mathbf{y}}, \mathbf{W}}\mathcal{L}(\hat{\mathbf{y}},\mathbf{W}) &= P(\mathbf{o}, \hat{\mathbf{y}}|\mathbf{x}; \mathbf{W}) \\
&= P(\mathbf{o}|\hat{\mathbf{y}})P(\hat{\mathbf{y}}|\mathbf{x}; \mathbf{W}),\\
\end{aligned}
\end{equation}
{where  $\mathbf{o}$ is  the observation of $\hat{\mathbf{y}}$, for example, the annotation can be seen 
	as an observation of $\hat{\mathbf{y}}$ from human annotator. 
	$P(\mathbf{o}|\hat{\mathbf{y}})$ is a prior of $\hat{\mathbf{y}}$ given the observation $\mathbf{o}$
	and $P(\hat{\mathbf{y}}|\mathbf{x}; \mathbf{W})$ is the likelihood. 
}

\subsection{{Prior model of `real' ground-truth}}
{To optimize Eq. (\ref{Eq:2}), an accurate prior model is important to regularize $\hat{\mathbf{y}}$ and reduce searching space. We assume that the $k$th landmark 
	$\hat{\mathbf{y}}^{k}$ is close to the $\mathbf{o}^{k}$, {which is the observation of $\hat{\mathbf{y}}$}. Thus, this prior  is modeled as Gaussian similarity over all   \{$\mathbf{o}^{k}, \hat{\mathbf{y}}^{k}$\} pairs:}
\begin{equation}\label{Eq:3}
\begin{split}
P(\mathbf{o}|\hat{\mathbf{y}})& \propto \prod_{k}\exp\Big(-\frac{\|\mathbf{o}^{k}-\hat{\mathbf{y}}^{k}\|^2}{2\sigma_1^2}\Big)\\
& =\exp\Big(-\sum_{k}\frac{\|\mathbf{o}^{k}-\hat{\mathbf{y}}^{k}\|^2}{2\sigma_1^2}\Big), \\
\end{split}
\end{equation}
{where  $\sigma_1$ can control the sensitivity to misalignment. To explain $\mathbf{o}^{k}$, we should know in advance that  
	our whole framework is iteratively optimized detailed in Section 4.4. $\mathbf{o}^{k}$ is initialized as the human annotation in the iteration, and will be updated by better observation with iterations.

	\begin{figure}[t]
		\begin{center}
			\includegraphics[width=.85\linewidth]{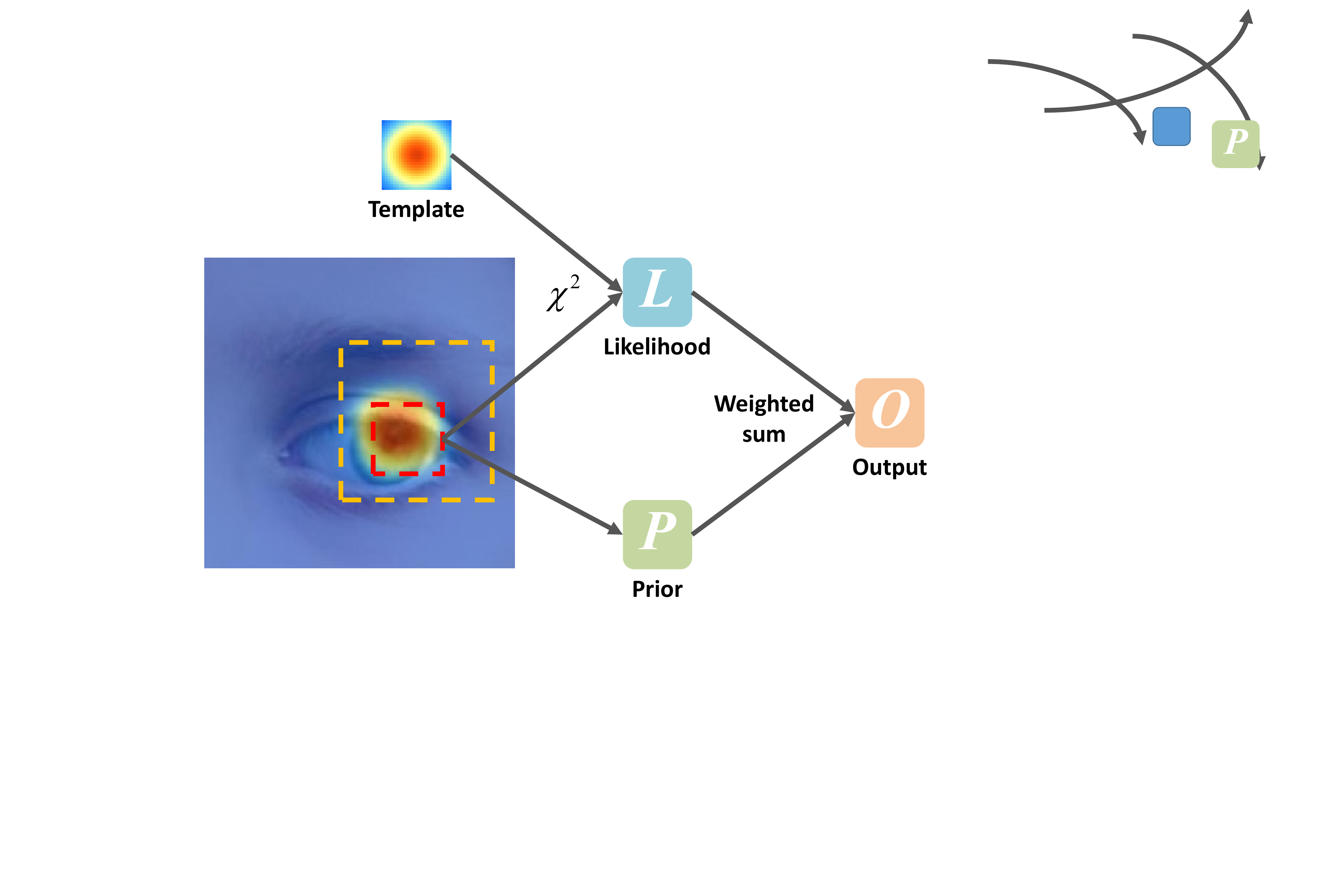}
		\end{center}
		\caption{{The search of  `real' ground-truth $\hat{\mathbf{y}}$. Yellow and red boxes represent the searching space $\mathcal{N}$ defined in Eq. (7) and the region corresponding to one candidate $\hat{\mathbf{y}}$, respectively. The weighted sum of likelihood and prior is computed as Eq. (8). 
				The search target is to find a position $\hat{\mathbf{y}}$ with the maximum output.}}
		\label{fig:heat_search}
	\end{figure}

	\subsection{Network likelihood model}
	{We now discuss the likelihood $P(\hat{\mathbf{y}}|\mathbf{x}; \mathbf{W})$ of Eq. (\ref{Eq:2}). 
		The point-wise joint probability can be represented by the confidence map, which can be modelled by the heatmap of the deep model. 
		Note that our hourglass architecture learns to predict heatmap consisting of a 2D Gaussian centered on the ground-truth 
		$\hat{\mathbf{y}}^{k}$. Thus, for any position $\mathbf{y}$, the more the heatmap  
		{region around}
		$\mathbf{y}$ follows a standard Gaussian, 
		the more the pixel at $\mathbf{y}$ is likely to be $\hat{\mathbf{y}}^{k}$. 
		Therefore, the likelihood  can be modeled as the distribution distance between the predicted heatmap (predicted distribution) and the standard Gaussian region (expected distribution).
		{In this work, we use {Pearson Chi-square test to evaluate the distance of these two distributions:} 
	}}
	
	\begin{equation}\label{Eq:4}
	\begin{split}
	\chi^{2}(\mathbf{y}|\mathbf{x}; \mathbf{W}) = \sum_{i}\frac{(\mathbf{E}_{i}-\Phi_{i}(\mathbf{y}|\mathbf{x}; \mathbf{W}))^{2}}{\mathbf{E}_{i}}
	\end{split}
	\end{equation}
	where $\mathbf{E}$ is a  standard Gaussian heatmap (distribution), which is a template representing the ideal response; 
	$i$ is the pixel index; $\Phi$ is a 
	cropped patch (of the same size as Gaussian template) from the {predicted heatmap} centered 
	on $\mathbf{y}$.
	Finally, the joint probability can also be modeled as a product of 
	Gaussian similarities maximized over all landmarks:
	
	\begin{equation}\label{Eq:5}
	\begin{split}
	P(\hat{\mathbf{y}}|\mathbf{x}; \mathbf{W}) = \exp\Big(-\sum_{k}\frac{\chi^{2}_{k}(\hat{\mathbf{y}}|\mathbf{x}; \mathbf{W})}{2\sigma_2^2}\Big)
	\end{split}
	\end{equation}
	where $k$ is the landmark index, $\sigma_2$ is the bandwidth of likelihood. 
	
	{ To keep the likelihood credible, {we first train a network with the human annotations.}
		Then in the likelihood, we can consider the trained network as a super annotator to guide the searching of the real ground-truth. 
		It results from the fact that 
		a well trained network is able to capture the statistical law of annotation noise from the whole training set, so that it can generate predictions with better semantic consistency.}

	\subsection{Optimization}
	Combining Eq. (\ref{Eq:2}), (\ref{Eq:3}) and (\ref{Eq:5}) and taking  log of the likelihood,
	we have:
	
	\begin{small}
		\begin{equation}\label{Eq:6}
		\begin{split}
		\log\mathcal{L}(\hat{\mathbf{y}},\mathbf{W})= \sum_{k}\Big(-\frac{\|\mathbf{o}^{k}-\hat{\mathbf{y}}^{k}\|^2}{2\sigma_{1}^2}  - 
		\frac{\chi^{2}(\hat{\mathbf{y}}|\mathbf{x}; \mathbf{W})}{2\sigma_{2}^2} \Big)
		\end{split}
		\end{equation}
	\end{small}
	{\paragraph{Reduce Searching Space} To optimize the latent {semantically} consistent `real' landmark $\hat{\mathbf{y}}^k$,  the prior Eq. (\ref{Eq:3}) indicates that  the latent `real' landmark is close to the observed landmark $\mathbf{o}^k$. }
	Therefore, we {reduce}
	the search space {of $\hat{\mathbf{y}}^k$} to a small patch centered on $\mathbf{o}^k$. Then, the optimization problem of Eq. (\ref{Eq:6}) can be re-written as:
	
	\begin{equation}\label{Eq:7}
	\begin{split}
	&\min_{\hat{\mathbf{y}},\mathbf{W}} -\log\mathcal{L}(\hat{\mathbf{y}},\mathbf{W})\\
	&~~~~\mathrm{s.t.}~~\hat{\mathbf{y}}^{k} \in \mathcal{N}(\mathbf{o}^{k})\\
	\end{split}
	\end{equation}
	where $\mathcal{N}(\mathbf{o}^{k})$ represents a region centered on ${\mathbf{o}}^{k}$. 
	
	\begin{figure*}
		\begin{center}
			\includegraphics[width=\textwidth]{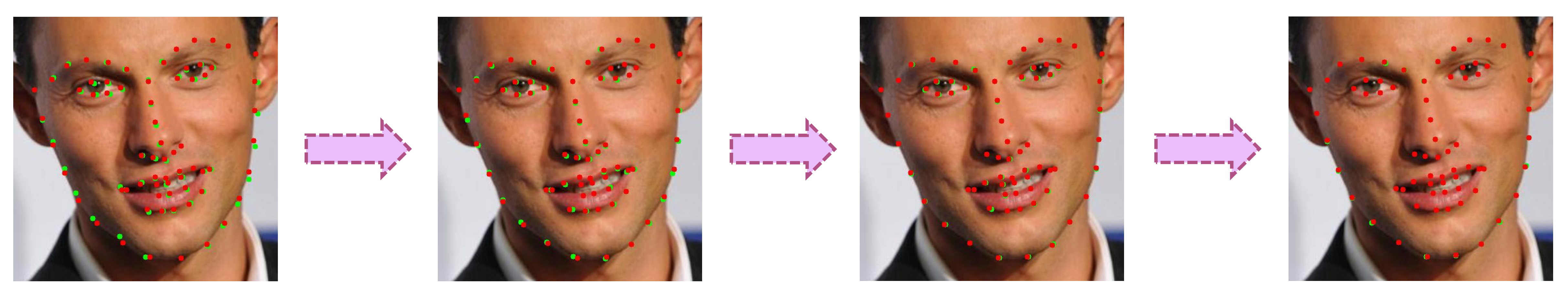}
		\end{center}
		\caption{Gradual convergence (one image represents one iteration) from the observation $\textbf{o}$ (i.e. $\hat{\mathbf{y}}$
			of the last iteration, green dots) to the estimate of real ground-truth $\hat{\mathbf{y}}$ (red dots). 
			For last image, the optimization converges because red and green dots are completely overlapped.}
		\label{fig:align}
	\end{figure*}

	\paragraph{Alternative Optimization}
	To optimize Eq. (\ref{Eq:7}), an alternative optimization strategy is applied. In each iteration, $\hat{\mathbf{y}}$ is 
	firstly searched with the network parameter $\mathbf{W}$  fixed. Then $\hat{\mathbf{y}}$ is fixed and $\mathbf{W}$ is 
	updated (landmark prediction network training) under the supervision of newly searched $\hat{\mathbf{y}}$.

	Step 1: When $\mathbf{W}$ is fixed, {to search the latent variable $\hat{\mathbf{y}}$}, the optimization becomes a constrained discrete optimization problem {for each landmark}:
	
	\begin{equation}\label{Eq:9}
	\begin{split}
	\min_{\hat{\mathbf{y}}^{k}}\Big(\frac{\|\mathbf{o}^{k}-\hat{\mathbf{y}}^{k}\|^2}{2\sigma_{1}^2}  +
	\frac{\chi^{2}(\hat{\mathbf{y}}^{k}|\mathbf{x}; \mathbf{W})}{2\sigma_{2}^2} \Big)
	\end{split}
	\end{equation}
	{where all the variables are known except $\hat{\mathbf{y}}^{k}$. We search $\hat{\mathbf{y}}^{k}$ by going through all the pixels in $\mathcal{N}(\mathbf{o}^{k})$ (a neighborhood area of $\mathbf{o}^{k}$ as shown in Fig.~\ref{fig:heat_search})
		and the one with minimal loss in Eq. (\ref{Eq:9}) is the solution. Since the searching space  $\mathcal{N}(\mathbf{o}^{k})$ is very small, i.e. 17 $\times$ 17 in this work for 256$\times$256 heatmap, the optimization is very efficient. 
		
		Note that in the prior part of Eq. (\ref{Eq:9}), $\mathbf{o}^{k}$ is the observation of $\hat{\mathbf{y}}^{k}$: In the 1st iteration, $\mathbf{o}^{k}$ is set to the
		human annotations which are the observation of human annotators;  
		From the 2nd iteration, $\mathbf{o}^{k}$  is set to $\hat{\mathbf{y}}^{k}_{t - 1}$ (where $t$ is 
		the iteration).
		Note that $\hat{\mathbf{y}}^{k}_{t - 1}$ is the estimated `real' ground-truth from the last iteration. 
		With the iterations, {$\hat{\mathbf{y}}^{k}_{t}$ is converging to the `real' ground-truth because 
			both the current observation $\mathbf{o}^{k}$ (i.e. $\hat{\mathbf{y}}^{k}_{t-1}$) and CNN prediction iteratively become more credible.}
		

		Step 2: When $\hat{\mathbf{y}}$ is fixed, the optimization becomes:
		\begin{equation}\label{Eq:8}
		\begin{split}
		\min_{\mathbf{W}}~\sum_{k}\frac{\chi^{2}(\hat{\mathbf{y}}^{k}|\mathbf{x}; \mathbf{W})}{2\sigma_{2}^2}
		\end{split}
		\end{equation}
		The optimization becomes a typical network training process under the supervision of $\hat{\mathbf{y}}$.
		Here 
		$\hat{\mathbf{y}}$ is set to the estimate of the latent `real' ground-truth obtained in Step 1.
		Figure~\ref{fig:align} shows an example of the gradual convergence from the observation $\textbf{o}$ ($\hat{\mathbf{y}}$
		of the last iteration) to the estimate of real ground-truth $\hat{\mathbf{y}}$. {The 
			optimization of $\hat{\mathbf{y}}$ in our semantic alignment can easily converge to a stable position, 
			which does not have hard convergence problem like the traditional landmark training as shown in Fig.~\ref{fig:analysis}b.}

		\begin{figure}
			\centering
			\subfigure[{ The use of GHCU for correcting some failed points.
			}]{
				\begin{minipage}[b]{0.45\textwidth}
					\includegraphics[width=\linewidth]{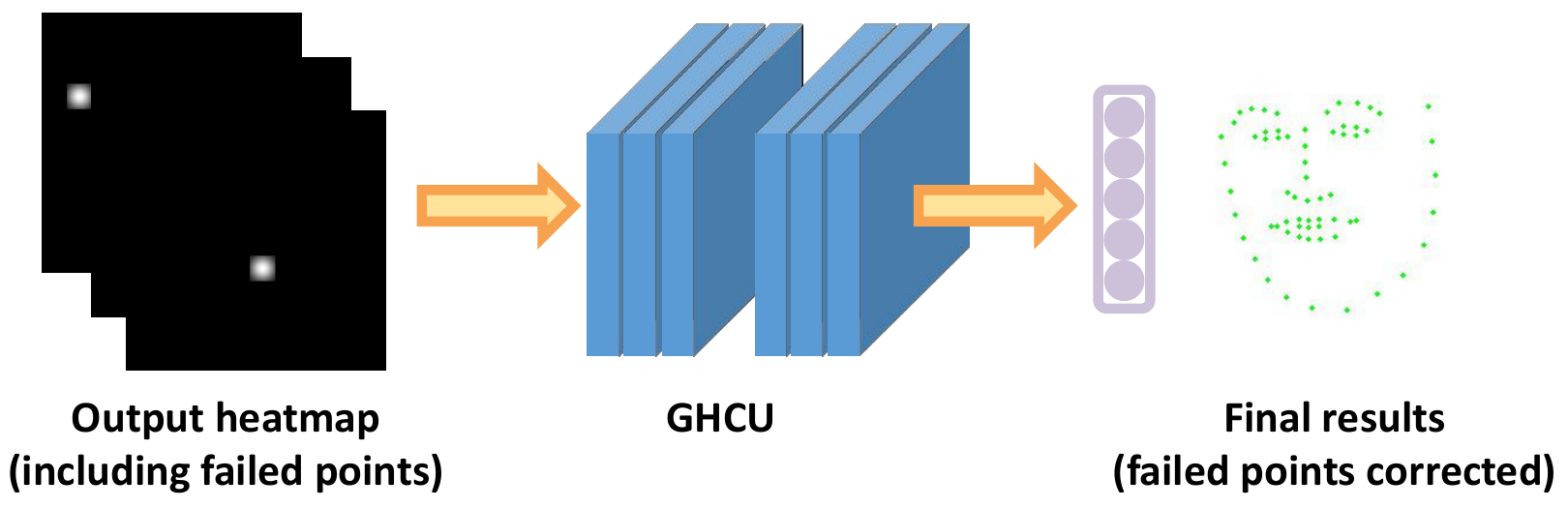} \\
				\end{minipage}
			}
			\subfigure[{ Correcting challenging points with GHCU on 300-VW.
			}]
			{
				\begin{minipage}[b]{0.45\textwidth}
					\includegraphics[width=\linewidth]{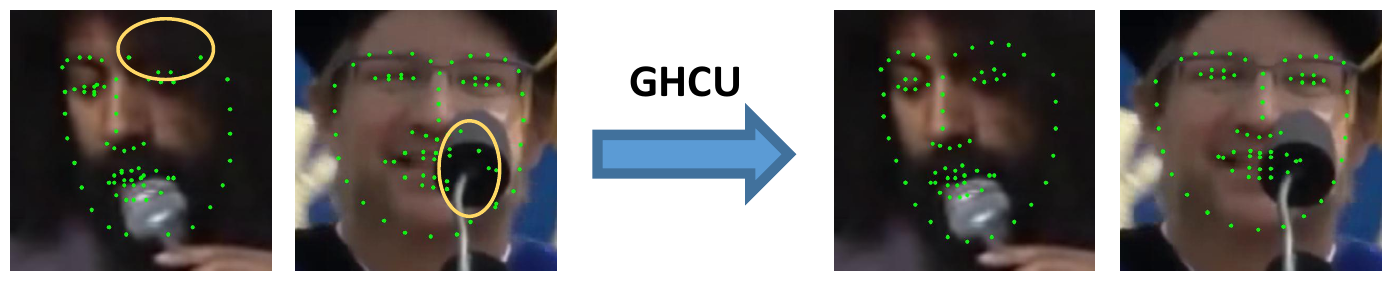}
				\end{minipage}
			}
			\caption{{Global Heatmap Correction Unit (GHCU)}}
			\label{fig:ghcu}
		\end{figure}
		
		\subsection{Global heatmap correction unit}
		
		Traditional heatmap based methods predict each landmark as an individual task without considering global face shape. 
		The prediction might fail when the model fits images of low-quality and occlusion as shown in Fig.~\ref{fig:ghcu}b.
		{The outliers such as occlusions} destroy the face shape and significantly reduce overall performance.
		
		Existing methods like local feature based CLM \cite{cristinacce2008automatic} and deep learning based 
		LGCN \cite{merget2018robust} apply a 2D shape PCA as their post-processing step to remove the outliers. However, PCA based method is 
		weak to model out-of-plane rotation and very slow (about 0.8 fps in LGCN \cite{merget2018robust}). In this work, 
		we propose a Global Heatmap Correction Unit (GHCU) to recover the outliers  efficiently. We 
		view the predicted heatmaps as input and directly regress 
		{the searched/optimized}
		$\hat{\mathbf{y}}$ through a 
		light weight CNN {as shown in Tab.~\ref{table_ghcu_arch}}. The GHCU implicitly learns the {whole} face shape constraint from the  training data and always gives facial-shape landmarks, 
		as shown in Fig.~\ref{fig:ghcu}. {Our experiments demonstrate the GHCU completes fitting with 
			the	speed 8 times faster than PCA on the same hardware platform and achieves higher accuracy than PCA.}
		
		
		\begin{table}[!t]
			\centering
			\caption{GHCU Architecture ($N$ is the number of the landmarks)}
			\vspace{3pt}
			\footnotesize
			\label{table_ghcu_arch}
			\begin{tabular}{c|c|c}
				\hline
				Layers  & Output size & GHCU \\
				\hline
				Conv1 & 128$\times$128 & [5$\times$5, 64], stride 2 \\
				Conv2 & 64$\times$64 & [3$\times$3, 64], stride 2 \\
				Conv3 & 32$\times$32 & [3$\times$3, 32], stride 2 \\
				Conv4 & 16$\times$16 & [3$\times$3, 32], stride 2 \\
				Conv5 & 8$\times$8 & [3$\times$3, 16], stride 2 \\
				Conv6 & 4$\times$4 & [3$\times$3, 16], stride 2 \\
				\hline
				FC1   & - & 256 \\
				FC2   & - &  $2N$ \\      
				\hline
			\end{tabular}
		\end{table} 
		
		\section{Experiments}
		\textbf{Datesets}. We conduct evaluation on three challenging
		datasets including image based 300W ~\cite{Sagonas2013A}, AFLW~\cite{koestinger2011annotated}, 
		and video based 300-VW ~\cite{shen2015first, tzimiropoulos2015project, chrysos2015offline}. 
		
		\emph{300W} ~\cite{Sagonas2013A} is a collection of multiple face datasets, including the LFPW~\cite{Belhumeur2011Localizing}, HELEN~\cite{Le2012Interactive}, AFW~\cite{Ramanan2012Face} and XM2VTS~\cite{Messer2000XM2VTS} which have 68 landmarks.
		The training set contains 3148 training samples, 689 testing samples which are further divided into the common and 
		challenging subsets.
		
		\emph{AFLW} ~\cite{koestinger2011annotated} is a very challenging dataset which has a wide range of pose variations in yaw
		($-90^\circ$ to $90^\circ$). In this work, we follow the AFLW-Full protocol~\cite{zhu2016unconstrained} which 
		ignores two landmarks of ears and use the remaining 19 landmarks. 
		
		\emph{300-VW} ~\cite{shen2015first, tzimiropoulos2015project, chrysos2015offline} is a large dataset for video-based face 
		alignment, which consists of 114 videos in various conditions.  Following ~\cite{shen2015first}, we utilized all images 
		from 300W and 50 sequences for training and the remaining 64 sequences for testing. The test set 
		consists of three categories: well-lit, mild unconstrained and challenging. 
		
		\textbf{Evaluation metric.} 
		To compare with existing popular methods, we conduct different evaluation metrics on different datasets.
		For 300W dataset, We follow the protocol in ~\cite{ren2016face} and use Normalized mean errors (NME) which 
		normalizes the error by the inter-pupil distance. For AFLW, we follow ~\cite{zhu2015face} to use face size as the normalizing factor. 
		For 300-VW dataset, we employed the standard normalized root mean squared error (RMSE) ~\cite{shen2015first} which normalizes the error by the outer eye corner distance. 
		
		\textbf{Implementation Details.}
		In our experiments, all the training and testing images are cropped and resized to 256$\times$256 according to the provided 
		bounding boxes. To perform data augmentation, we 
		randomly sample the angle of rotation and the bounding box scale from Gaussian distribution. We use a four-stage 
		stacked hourglass network ~\cite{newell2016stacked} as our backbone which is trained by the optimizer RMSprop. 
		{As described in Section 4}, our algorithm comprises two parts: network training and real ground-truth searching, which are alternatively optimized. Specifically, at each epoch, we first search the real ground-truth $\hat{\mathbf{y}}$ and 
		then use $\hat{\mathbf{y}}$ to supervise the network training. 
		{When training the roughly converged model with human annotations, the initial learning rate is $2.5 \times 10^{-4}$ which is 
			decayed to $2.5 \times 10^{-6}$ after 120 epochs. When training with Semantic Alignment from the beginning of the aforementioned 
			roughly converged model, the initial learning rate is $2.5 \times 10^{-6}$ and is divided by 5, 2 and 2 at epoch 30, 60 and 
			90 respectively.
			During semantic alignment,  we search the latent variable $\hat{\mathbf{y}}$ from a 17$\times$17 region centered at the current observation 
			point $\mathbf{o}$, and we crop a no larger than 25$\times$25 patch from the predicted heatmap around current position 
			for Pearson Chi-square test in Eq. (\ref{Eq:4}).}
		{We set batch size to 10} for network training. For GHCU, the network architecture is shown in Tab.~\ref{table_ghcu_arch}. 
		All our models are trained with PyTorch ~\cite{paszke2017automatic} on 2 Titan X GPUs.

		\begin{table}[!thp]
			\centering
			\caption{Comparisons with state of the art on 300W dataset. The error (NME) is normalized by the inter-pupil distance.}
			\vspace{3pt}
			\footnotesize
			\label{table_300w}
			\begin{tabular}{l|ccc}
				\hline
				\diagbox{Method}{subset}   & Com. & Challenge & Full \\
				\hline
				SDM~\cite{Xiong2013Supervised}    & 5.60 & 15.40 & 7.52 \\
				CFSS~\cite{zhu2015face}   & 4.73 & 9.98  & 5.76 \\
				TCDCN~\cite{Zhang2014Facial}  & 4.80  & 8.60  & 5.54 \\
				LBF~\cite{ren2016face}    & 4.95 & 11.98 & 6.32\\
				3DDFA (CVPR16)~\cite{Zhu2016Face} & 6.15 & 10.59 & 7.01 \\
				3DDFA + SDM & 5.53 & 9.56 & 6.31 \\
				RAR (ECCV16)~\cite{xiao2016robust} & 4.12 & 8.35 & 4.94 \\
				TR-DRN (CVPR17)~\cite{lv2017deep} & 4.36 & 7.56  & 4.99 \\
				Wing (CVPR18)~\cite{feng2017wing} & \textbf{3.27} & 7.18  & 4.04 \\
				LAB (CVPR18)~\cite{wu2018look} & 3.42 & 6.98  & 4.12 \\
				SBR (CVPR18)~\cite{dong2018supervision} & 3.28 & 7.58  & 4.10 \\
				PCD-CNN (CVPR18)~\cite{kumar2018disentangling} & 3.67 & 7.62  & 4.44 \\
				DCFE (ECCV18)~\cite{valle2018deeply} & 3.83 & 7.54  & 4.55 \\
				\hline
				HGs & 4.43  & 7.56  & 5.04  \\
				\textbf{HGs + SA} & 3.75 & 6.90  & 4.37  \\
				\textbf{HGs + SA + GHCU} & 3.74 & 6.87  & 4.35  \\
				HGs + Norm & 3.95  & 6.51  & 4.45  \\
				\textbf{HGs + SA + Norm} &  3.46 & 6.38  & 4.03  \\
				\textbf{HGs + SA + Norm + GHCU} &  3.45 & \textbf{6.38}  & \textbf{4.02}  \\
				\hline
			\end{tabular}
		\end{table}
		
		\subsection{Comparison experiment}
		
		{\textbf{300W.}  We compare our approach against the state-of-the-art methods on 300W in Tab.~\ref{table_300w}. 
			{The baseline (HGs in Tab.~\ref{table_300w}) uses the hourglass  architecture  with human annotations, which is actually the traditional landmark detector training.} From Tab.~\ref{table_300w},
			we can see that HGs with our Semantic Alignment (HGs + SA) greatly outperform hourglass (HGs) only, 4.37\% vs 5.04\% in terms of NME on Full set, showing the great effectiveness of our Semantic Alignment (SA). By adding GHCU, we can see that HGs+SA+GHCU slightly outperforms the HGs+SA. {The improvement is not significant because the images of 300W are of high resolution, while GHCU works particularly well for images of low resolution and occlusions verified in the following evaluations.}
			Following {~\cite{feng2017wing} and \cite{Yang2017Stacked}} which normalize the in-plane-rotation {by training a preprocessing network}, we conduct this normalization (HGs+SA+GHCU+Norm)  and achieve state of the art performance on Challenge set and Full set: 6.38\% and 4.02\%. In particular, on Challenge set, we significantly outperform the state of the art method: 6.38\% (HGs+SA+GHCU+Norm) vs 6.98\% (LAB), meaning that our method is particularly effective on challenging scenarios. 
			

			\textbf{AFLW.}
			300W has 68 facial points which contain many weak semantic landmarks (e.g. those on face contours). In comparison, AFLW has only 19 points, most of which are strong semantic landmarks. Since our SA is particularly effective on weak semantic points, we conduct experiments on AFLW to verify whether SA generalizes well to the point set, most of which are strong semantic points.  For fair comparison, we do not compare methods using additional outside training data, e.g. LAB ~\cite{wu2018look} used additional boundary information from 
			outside database.
			As shown in Tab.~\ref{table_aflw}, HGs+SA  outperforms HGs, 1.62\% vs 1.95\%. It means that even though corner points are easily to be recognized, there is still random error in annotation, which can be corrected by SA. {It is also observed that HGs+SA+GHCU works better than 
				HGs+SA.} 
			

			\begin{table}[!thp]
				\centering
				\caption{Comparison with state of the art on AFLW dataset. The error (NME) is normalized by the face bounding box size.}
				\vspace{3pt}
				\footnotesize
				\label{table_aflw}
				\begin{tabular}{l|cc}
					\hline
					Method   & AFLW-Full (\%)\\
					\hline
					LBF~\cite{ren2016face} & 4.25 \\
					CFSS~\cite{zhu2015face} & 3.92 \\
					CCL (CVPR16)~\cite{zhu2016unconstrained} & 2.72 \\
					TSR (CVPR17)~\cite{lv2017deep} & 2.17 \\
					DCFE (ECCV18)~\cite{valle2018deeply} & 2.17 \\
					SBR (CVPR18)~\cite{dong2018supervision} & 2.14 \\
					DSRN (CVPR18)~\cite{miao2018direct} & 1.86 \\
					Wing (CVPR18)~\cite{feng2017wing} & 1.65 \\
					\hline
					HGs & 1.95 \\
					\textbf{HGs + SA} & 1.62 \\
					\textbf{HGs + SA + GHCU} & \textbf{1.60} \\
					\hline
				\end{tabular}
			\end{table}

			\textbf{300-VW.} 
			Unlike the image-based databases 300W and AFLW, 300-VW is video-based database, which is more challenging because the frame is of low resolution and with strong occlusions. The subset Category 3 is the most challenging one. From Tab.~\ref{table_300vw}, 
			{we can see that HGs + SA greatly outperforms HGs in each of these three test sets. Furthermore, }
			compared with HGs + SA, HGs + SA + GHCU reduce the error rate (RMSE) by 18\% {on Category 3 test set}, meaning that GHCU is very effective for video-based challenges such as low resolution and occlusions because GHCU 
			considers
			the global face shape as constraint, being robust to such challenging factors.
			

			\begin{table}[!thp]
				\centering
				\caption{Comparison with state of the art on 300-VW dataset. The error (RMSE) is normalized by the inter-ocular distance.}
				\vspace{3pt}
				\footnotesize
				\label{table_300vw}
				\begin{tabular}{l|ccc}
					\hline
					Method   & Category 1 & Category 2 & Category 3 \\
					\hline
					SDM~\cite{Xiong2013Supervised}   & 7.41 & 6.18  \\
					CFSS~\cite{zhu2015face}   & 7.68 & 6.42 & 13.67 \\
					TCDCN~\cite{zhang2016learning}   & 7.66 & 6.77 & 14.98 \\
					TSTN~\cite{liu2017two}   & 5.36 & 4.51 & 12.84 \\
					DSRN (CVPR18)~\cite{miao2018direct} & 5.33 & 4.92 & 8.85 \\
					\hline
					HGs & 4.32 & 3.83 & 9.91 \\
					\textbf{HGs + SA} & 4.06 & 3.58 & 9.19 \\
					\textbf{HGs + SA + GHCU} & \textbf{3.85} & \textbf{3.46} & \textbf{7.51} \\
					\hline
				\end{tabular}
			\end{table}

			\subsection{Self evaluations}
			
			\textbf{Balance of prior and likelihood} 
			As shown in Eq. (\ref{Eq:6}), the `real' ground-truth is optimized using two parts:
			prior and likelihood, where  $\sigma_{1}$ and $\sigma_{2}$ determine the importance of these two 
			parts. Thus, we can use one parameter $\sigma_{2}^2 / \sigma_{1}^2$ to estimate this importance weighting.
			We evaluate different values of $\sigma_{2}^2 / \sigma_{1}^2$ in Tab.~\ref{table_sigma}. 
			Clearly, the performance of $\sigma_{2}^2 / \sigma_{1}^2 = 0$ (removing Semantic Alignment and  using human annotations only) is worst, showing the importance of the proposed Semantic Alignment.
			We find that $\sigma_{2}^2 / \sigma_{1}^2 = 0.1$ achieves the best performance, meaning that the model relies much more (10 times) on prior than likelihood to achieve the best trade-off. 

			\begin{table}[!thp]
				\centering
				\caption{The effect of the ratio   $\sigma_{2}^2 / \sigma_{1}^2$ in Eq. (\ref{Eq:9}) on 300W.}
				\vspace{3pt}
				\footnotesize
				\label{table_sigma}
				\begin{tabular}{l|ccccccc}
					\hline
					$\sigma_{2}^2 / \sigma_{1}^2 $   & 0 & 0.01 & 0.05 & 0.1 & 0.3 & 0.5 & 1 \\
					\hline
					NME (\%) & 4.99 & 4.79 & 4.40 & \textbf{4.37} & 4.46 & 4.54 & 4.68 \\
					\hline
				\end{tabular}
			\end{table}
			
			\textbf{Template size.} As discussed in the Section 3, 
			for a position $\mathbf{y}$, the similarity between the heatmap region around it and standard Gaussian template 
			is
			closely related to the detection confidence. Therefore, the size of the Gaussian template, which is used 
			to {measure the network confidence} in Eq. (\ref{Eq:5}), can affect the final results. 
			Table~\ref{table_region} reports the results under different template sizes using the model HGs+SA. 
			Too small size (size=1)  means 
			that the heatmap value is directly used to model the likelihood  instead of Chi-square test.  Not surprisingly, the performance with size=1 is not promising. Large size (size=25) introduces more useless information, degrading the performance. 
			In our experiment, {we find size=15 for AFLW and size=19 for 300W} can achieve the best result. 
			

			\begin{table}[!thp]
				\centering
				\caption{The effects of template size 
					on 300W  and AFLW test sets.}
				\vspace{3pt}
				\footnotesize
				\label{table_region}
				\begin{tabular}{l|cccccc}
					\hline
					template size    & 1 & 7 & 11 & 15 & 19 & 25 \\
					\hline
					300W Full(\%) & 4.76 & 4.72 & 4.61 & 4.53 & \textbf{4.37} & 4.43 \\
					\hline
					AFLW Full (\%) & 1.89 & 1.80 & 1.72 & \textbf{1.62} & 1.66 & 1.70 \\
					\hline
				\end{tabular}
			\end{table}
			
			\textbf{Analysis of the training of semantic alignment.} 
			To verify the effectiveness of Semantic Alignment, we train a baseline network using hourglass 
			under the supervision of
			human annotation to converge. 
			Use this roughly converged baseline, we continue training using
			3 strategies as shown in Fig.~\ref{fig:loss_compare} and~\ref{fig:nme_compare}: baseline, SA w/o update (always using human annotation as the observation
			, see Eq. (\ref{Eq:6})) and SA (the observation is {iteratively} updated). 
			Fig.~\ref{fig:loss_compare} and~\ref{fig:nme_compare} {visualize the changes of training loss and NME on test set against the training epochs, respectively.}
			{We can see that the baseline curve in Fig.~\ref{fig:loss_compare} and ~\ref{fig:nme_compare} do not decrease because of the `semantic ambiguity'. By introducing SA, the training loss and test NME steadily drop. Obviously, SA reduces the random optimizing directions and helps the roughly converged network to further improve the detection accuracy.
			}
			
			{We also evaluate the condition that uses semantic alignment without updating the observation $\mathbf{o}$ (`SA w/o update' in Fig. ~\ref{fig:loss_compare} and ~\ref{fig:nme_compare}). It means  $\mathbf{o}$ is always set to the human annotations. 
				We can see that 
				the curve of `SA w/o update' can be further optimized but quickly trapped into local optima, leading to
				worse performance than SA. We assume that the immutable observation $\mathbf{o}$ reduces the capacity of searching `real' 
				ground-truth $\hat{\mathbf{y}}$. }

		}

		\begin{figure}[!thp]
			\begin{center}
				\includegraphics[width=\linewidth]{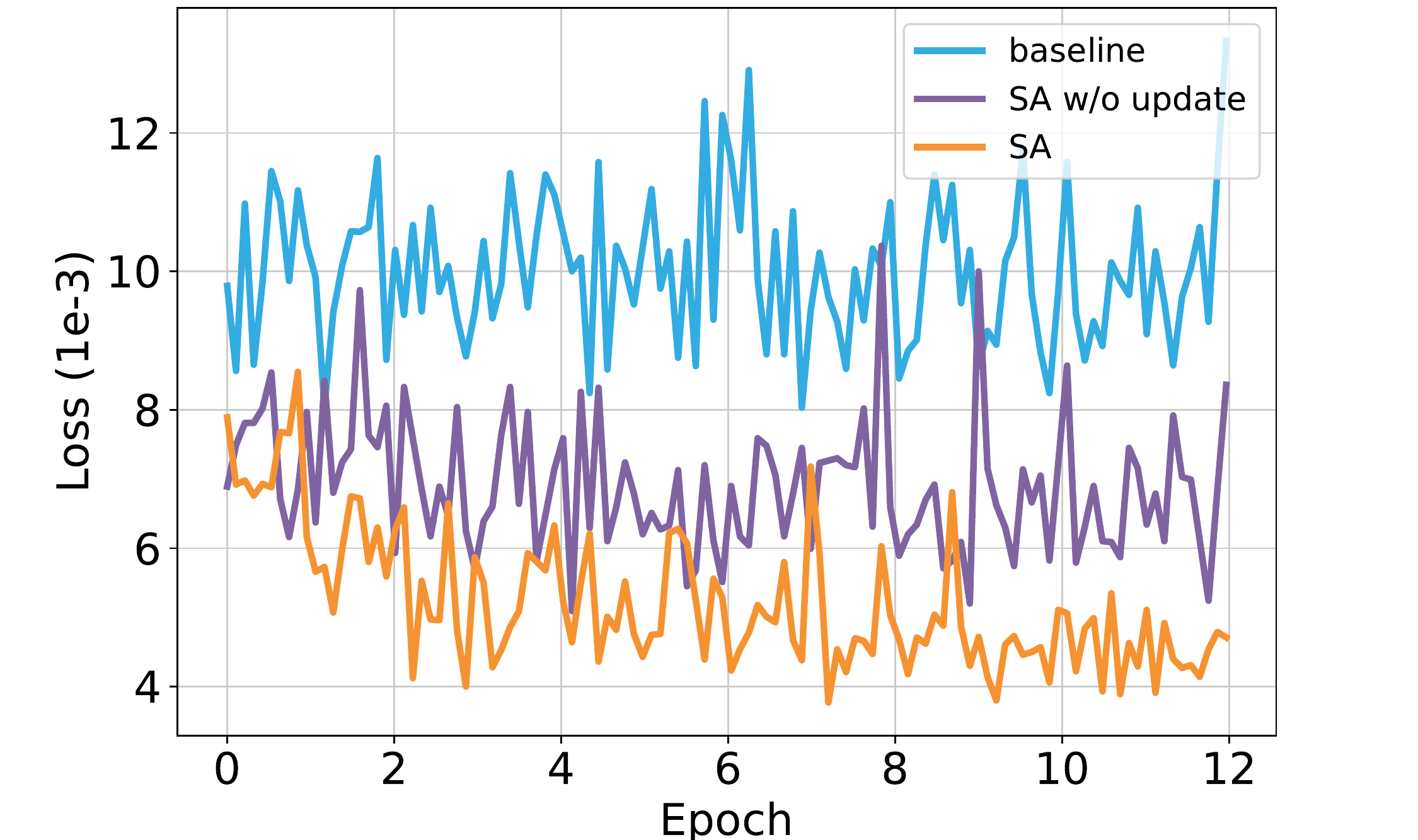}
			\end{center}
			\caption{Training loss of the baseline, 
				Semantic Alignment without updating observation (SA w/o update) and Semantic Alignment (SA). The training starts at a roughly converged model (trained using human annotations only) using 300W training set.}
			\label{fig:loss_compare}
		\end{figure}
		

		\begin{figure}[!thp]
			\begin{center}
				\includegraphics[width=\linewidth]{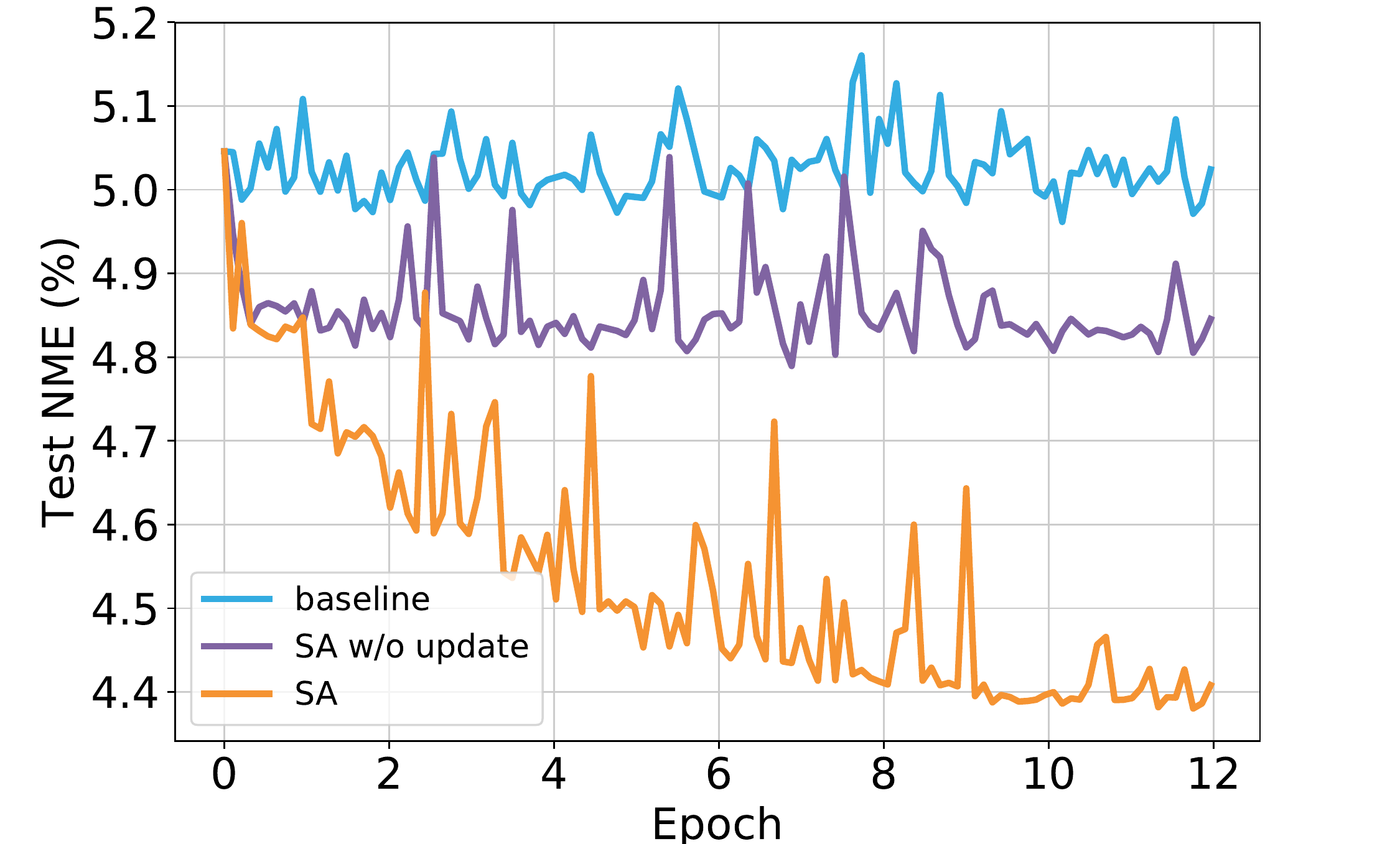}
			\end{center}
			\caption{NME of the baseline, 
				Semantic Alignment without updating observation (SA w/o update) and Semantic Alignment (SA). The training starts at a roughly converged model (trained using human annotations only) on 300W full test set.}
			\label{fig:nme_compare}
		\end{figure}

		\textbf{The update of Semantic Alignment.}
		Under Semantic Alignment framework, all the training labels are updated after each epoch. {To explore the effects of the number 
			of epochs 
			on model convergence, }we train different 
		models by stopping  semantic alignment at different epochs.
		In 	Fig~\ref{fig:align_num1}, it is  observed that the final performance keeps 
		improving
		with the times of semantic alignment, 
		{which demonstrates that the improvement is highly positive related to the quality of the learned $\hat{\mathbf{y}}$.
		}
		From our experiment, 10 epochs of semantic alignment are enough for our data sets.
		
		\begin{figure}[!thp]
			\begin{center}
				\includegraphics[width=\linewidth]{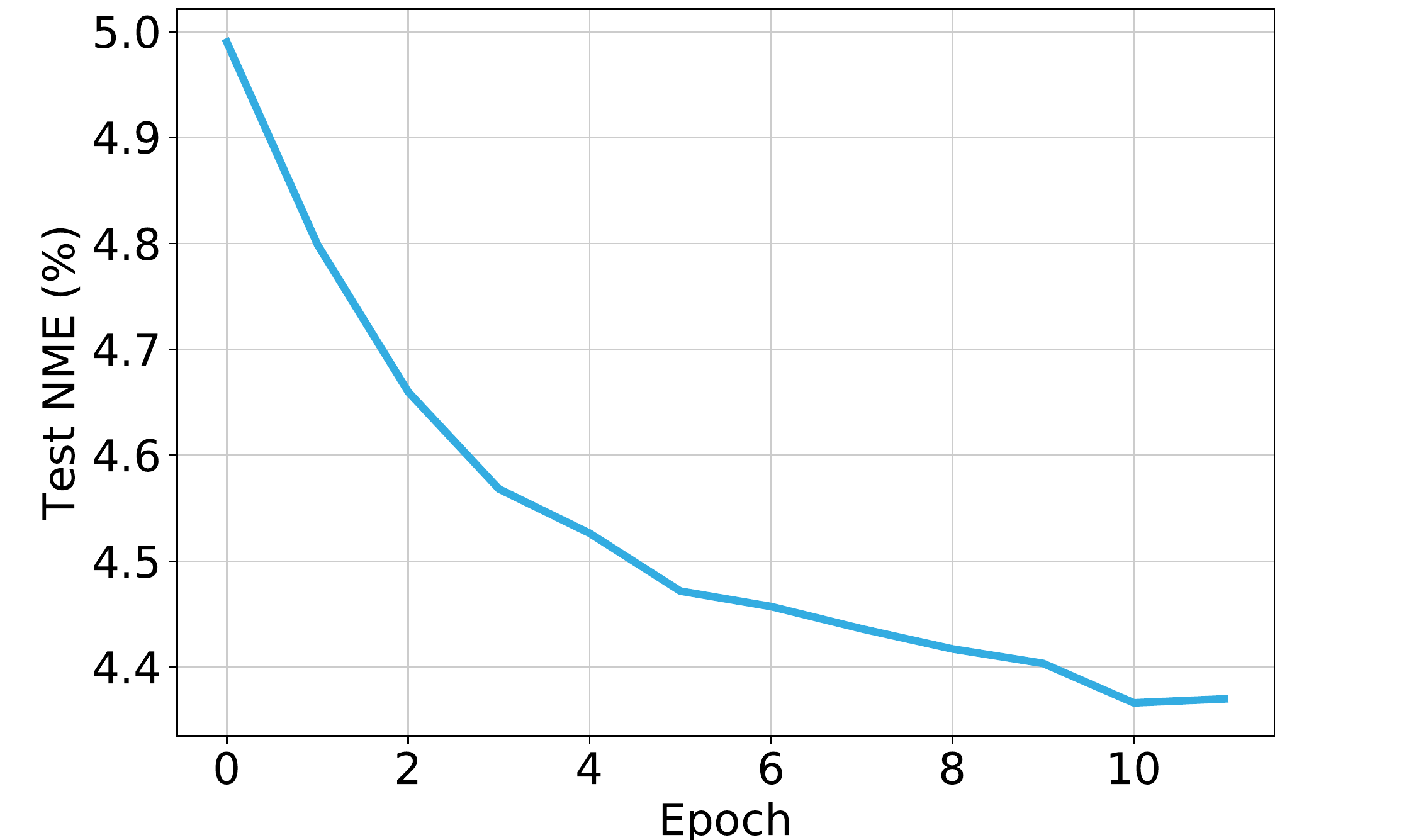}
			\end{center}
			\caption{NME vs Semantic Alignment update epochs on 300W full test set}
			\label{fig:align_num1}
		\end{figure}

		\textbf{Quality of the searched `real' ground-truth.}
		{One important assumption of this work is that there exist `real' ground-truths which are better than the human annotations. To verify this, we train two networks which are supervised by the human annotations 
			provided by public database and the searched `real' ground-truth, respectively. These two detectors are a 
			Hourglass model (HGs) and a ResNet ~\cite{he2016deep} model which directly regresses the landmark 
			coordinates as \cite{feng2017wing}. As shown in Tab.~\ref{table_label_qa}, we can see that on both models the `real' ground-truth (SA) outperforms 
			the human annotations (HA). Clearly, our learned labels are better than the human annotations, verifying our assumption that the semantic alignment can find the semantic consistent ground-truths. 
		}


		\begin{table}[!thp]
			\centering
			\caption{
				The comparison of the labels searched by our Semantic Alignment (SA) and human annotations (HA)  on 300w-full set
			}
			\vspace{3pt}
			\footnotesize
			\label{table_label_qa}
			\begin{tabular}{l|cccc}
				\hline
				Method   & HGs (HA) & \textbf{HGs (SA) } & Reg (HA) & \textbf{Reg (SA)} \\
				\hline
				NME (\%) & 5.04 & \textbf{4.37} & 5.49 & \textbf{5.12} \\
				\hline
			\end{tabular}
		\end{table}
		

		\textbf{Global heatmap correction unit.} 
		The 2D shape PCA can well keep the face constraint and can be conducted as 
		a post-processing step to enhance the performance of heatmap based methods, like CLM \cite{cristinacce2008automatic} and 
		most recently LGCN \cite{merget2018robust}. We apply the powerful PCA refinement method in LGCN and compare it with our 
		GHCU. We evaluate on 300-VW where the occlusion and low-quality are particularly challenging. 
		As shown in Tab.~\ref{table_pca_compare}, our CNN based GHCU outperforms PCA based method in terms of both accuracy and efficiency.

		\begin{table}[hpt]
			\centering
			\caption{The comparison of GHCU with traditional PCA-based refinement on 300-VW database.}
			\vspace{3pt}
			\footnotesize
			\label{table_pca_compare}
			\begin{tabular}{l|cccc}
				\hline
				Method   & Category 1 & Category 2 & Category 3 & CPU Time (ms)  \\
				\hline
				Baseline   & 4.06 & 3.58  & 9.19 & -\\
				PCA~\cite{merget2018robust}   & 3.99 & \textbf{3.26}  & 7.69 & 1219 \\
				\textbf{GHCU}   & \textbf{3.85} & 3.46  & \textbf{7.51} & 149 \\
				\hline
			\end{tabular}
		\end{table}

		\textbf{Ablation study.} 
		To verify the effectiveness of different components in our framework, we conduct this ablation study on 300-VW. 
		For a fair comparison, all the experiments use the same 
		parameter settings.  
		As shown in Tab.~\ref{table_ablation}, 
		Semantic alignment can consistently improve  the performance on all subset sets,  demonstrating the strong generalization capacity 
		of SA. GHCU is 
		more effective on the challenge data set (Category 3): 8.15\% vs 9.91\%;  
		Combining SA and GHCU works better than single of them, showing the complementary of these two mechanisms. 
		
		\begin{table}[!thp]
			\centering
			\caption{Effectiveness of SA and  GHCU tested on 300-VW.}
			\vspace{3pt}
			\footnotesize
			\label{table_ablation}
			\begin{tabular}{l|cccc}
				\hline
				Semantic Alignment (SA) & \checkmark &  & \checkmark &  \\
				GHCU & \checkmark & \checkmark &  &  \\
				\hline
				Category 1   & \textbf{3.85} & 4.03  & 4.06 & 4.32 \\
				Category 2   & \textbf{3.46} & 3.66  & 3.58 & 3.83 \\
				Category 3   & \textbf{7.51} & 8.15  & 9.19 & 9.91 \\
				\hline
			\end{tabular}
		\end{table}

		\section{Conclusion}
		In this paper, we first analyze the semantic ambiguity of facial landmarks and show that the potential
		random noises of landmark annotations  can  degrade the performance considerably. To address this issue, we propose a 
		a novel latent variable optimization strategy to find the
		{semantically} consistent annotations and alleviate random noises during 
		training stage. Extensive experiments demonstrated that our method effectively improves the landmark detection accuracy on different data sets.

{\small
\bibliographystyle{ieee_fullname}
\bibliography{egbib}
}

\end{document}